\documentclass[11pt]{article}
\usepackage{acl2012}
\usepackage{times}
\usepackage{latexsym}
\usepackage{amsmath,amssymb}
\usepackage{multirow}
\usepackage{url}
\usepackage{color}
\usepackage{graphicx}
\usepackage{bbm}
\usepackage{xspace}
\usepackage{wasysym}
\usepackage[titletoc,title]{appendix}

\usepackage{latexsym}
\usepackage{graphicx}
\usepackage{algorithmic}
\usepackage{float}
\usepackage{mathtools}
\usepackage{array}
\usepackage{graphics}
\usepackage{comment}
\usepackage{caption}
\usepackage{etoolbox}

\newcommand\norm[1]{\left\lVert#1\right\rVert}

\newcommand\T{\rule{0pt}{2.2ex}}

\definecolor{mbgreen}{RGB}{34,189,64}

\newcommand{\kpgcomment}[1]{\textcolor{blue}{\bf \small [ #1 --KPG]}}

\newcommand{\skipgram}{skip-gram\xspace}

\newcommand{\annoppdb}{Annotated-PPDB\xspace}

\newcommand{\paragram}{\textsc{paragram}\xspace}
\newcommand{\paragramsubs}[1]{\textsc{paragram}$_{#1}$}
\newcommand{\annoppdbthreek}{Annotated-PPDB-3K\xspace}
\newcommand{\mlpara}{ML-Paraphrase\xspace}
\newcommand{\wsall}{WS353\xspace}
\newcommand{\wssim}{WS-S\xspace}
\newcommand{\wsrel}{WS-R\xspace}
\newcommand{\simlex}{SL999\xspace}
\newcommand{\singleast}{$\phantom{^{\ast}}$}
\newcommand{\doubleast}{$\phantom{^{\ast\dagger}}$}
\newcommand{\tripleast}{$\phantom{^{\ast\dagger\ddagger}}$}
\newcolumntype{C}[1]{>{\centering\arraybackslash}p{#1}}

\newcounter{Lcount1}
\newcommand{\squishlist}{
 \begin{list}{\arabic{Lcount1}. }
  { \usecounter{Lcount1}
\setlength{\itemsep}{1pt}
\setlength{\parsep}{2pt}
\setlength{\topsep}{1pt}
\setlength{\partopsep}{0pt}
\setlength{\leftmargin}{0.5em}
\setlength{\labelwidth}{0.4em}
\setlength{\labelsep}{0.1em} } }

\newcounter{Lcount2}
\newcommand{\squishlisttwo}{
\begin{list}{\alph{Lcount2}. }
{ \usecounter{Lcount2}
\setlength{\itemsep}{1pt}
\setlength{\parsep}{0pt}
\setlength{\topsep}{0pt}
\setlength{\partopsep}{0pt}
\setlength{\leftmargin}{1.0em}
\setlength{\labelwidth}{0.4em}
\setlength{\labelsep}{0.1em} } }
\newcommand{\squishend}{
\end{list} }

\def\secref#1{\S\ref{#1}}

\DeclareMathOperator*{\argmax}{argmax}

\makeatletter
\newcommand{\@BIBLABEL}{\@emptybiblabel}
\newcommand{\@emptybiblabel}[1]{}
\makeatother
\title{From Paraphrase Database to Compositional Paraphrase Model and Back}

\author{
   John Wieting$^{\ast}$
\ \ \ \  Mohit Bansal$^{\dag}$ 
\ \ \ \  Kevin Gimpel$^{\dag}$
\ \ \ \  Karen Livescu$^{\dag}$
\ \ \ \  Dan Roth$^{\ast}$
\\
$^{\ast}$University of Illinois at Urbana-Champaign, Urbana, IL, 61801, USA\\
\tt{\{wieting2,danr\}@illinois.edu}\\
$^{\dag}$Toyota Technological Institute at Chicago, Chicago, IL, 60637, USA\\
\tt{\{mbansal,kgimpel,klivescu\}@ttic.edu}
}

\begin{document}
\maketitle
\begin{abstract}

The Paraphrase Database (PPDB; Ganitkevitch et al., 2013)\nocite{GanitkevitchDC13} is an extensive semantic resource,
consisting of a list of phrase pairs with (heuristic) confidence estimates. 
However, it is still unclear how it can best be used, due to the heuristic nature of the confidences and its necessarily incomplete coverage.
We propose models to leverage the phrase pairs from the PPDB to build parametric 
paraphrase models that score paraphrase pairs more accurately than the PPDB's internal scores while simultaneously
improving its coverage.
They allow for learning phrase embeddings as well as improved word embeddings.
Moreover, we 
introduce two new, manually annotated datasets to evaluate short-phrase paraphrasing models. 
Using our paraphrase model trained using PPDB, we achieve state-of-the-art results on standard word and bigram similarity tasks and beat strong baselines on our new short phrase paraphrase tasks.\footnote{We release our datasets, code, and trained models on the authors' websites.}$^{,}$\footnote{This version differs from the previous one with the inclusion of Appendix A, which contains details about new higher dimensional embeddings we have released. These embeddings achieve human-level performance on \simlex and \wsall.}

\end{abstract}

\section{Introduction}
Paraphrase detection\footnote{See \newcite{androutsopoulos2010survey} for a survey on approaches for detecting paraphrases.}  is the task of analyzing two segments of text and determining if they have the same meaning despite differences in structure and wording. It is useful for a variety of NLP tasks like question answering~\cite{rinaldi-03,fader-13}, semantic parsing~\cite{berant2014semantic}, textual entailment \cite{bosma-06}, and machine translation~\cite{marton-09}.

One component of many such systems is a paraphrase table containing pairs of text snippets, usually automatically generated, that have the same meaning. The most recent work in this area is the Paraphrase Database (PPDB; Ganitkevitch et al., 2013), \nocite{GanitkevitchDC13} a collection of confidence-rated paraphrases created using the 
pivoting technique of \newcite{bannard2005paraphrasing} over large parallel 
corpora. 
The PPDB is a massive resource, containing 220 million paraphrase pairs. It captures many short paraphrases that would be difficult to obtain using any other 
resource. For example, the pair \{{\it we must do our utmost}, {\it we must make every effort}\} has little lexical overlap but is present in PPDB. The PPDB has recently been used for monolingual alignment~\cite{yao2013semi}, for predicting sentence similarity~\cite{bjerva2014meaning}, and to improve the coverage of FrameNet~\cite{rastogi2014augmenting}.

Though already effective for multiple NLP tasks, we note some drawbacks of PPDB. 
The first 
is lack of coverage: 
to use the PPDB to compare two phrases, 
both must be in the database. 
The second is that PPDB is a nonparametric paraphrase model; the number of parameters (phrase pairs) grows with the size of the dataset used to build it. In practice, 
it can become unwieldy to work with 
as the size of the database increases. A third concern is that the confidence estimates in PPDB are a heuristic combination of features, and their quality is unclear.

We address these issues in this work by introducing 
ways to use PPDB to construct parametric paraphrase models. First we show that initial skip-gram word vectors \cite{mikolov2013efficient}  
can be fine-tuned for the paraphrase task by training on word pairs from PPDB. We call them {\bf \paragram} word vectors. 
We find additive composition of \paragram vectors to be 
a simple but effective way to embed phrases 
for short-phrase paraphrase tasks. 
We 
find improved performance 
by training a recursive neural network (RNN; Socher et al., 2010)\nocite{socher2010learning} directly on phrase pairs from PPDB.

We show that our resulting word and phrase 
representations are effective on a wide variety of 
tasks, including two new 
datasets that we introduce. 
The first, 
\annoppdb, contains pairs from PPDB that were 
scored by human annotators. 
It can be used to evaluate paraphrase models for short phrases. 
We use it to show that 
the phrase embeddings produced by our methods are significantly more indicative of paraphrasability than the original heuristic scoring used by \newcite{GanitkevitchDC13}. 
Thus we use the power of PPDB to improve its contents.

Our second dataset, \mlpara, is a re-annotation of the bigram similarity corpus from \newcite{Mitchell:Lapata:2010}. The task  
was 
originally 
developed 
to measure semantic similarity of bigrams, but some 
annotations are not congruent with the functional similarity central to paraphrase relationships. 
Our re-annotation can be used to assess paraphrasing capability of bigram compositional models.

\noindent In summary, we make the following contributions: 

\vspace{2pt}
\noindent\textbf{Provide new \paragram word vectors}, 
learned using PPDB, that achieve state-of-the-art performance on the SimLex-999 lexical similarity task~\cite{HillRK14} and  
lead to improved performance in 
sentiment analysis. 

\vspace{2pt}
\noindent\textbf{Provide ways to use PPDB to embed phrases.} We compare additive and RNN composition of \paragram vectors. 
Both can improve PPDB 
by re-ranking the paraphrases in PPDB to 
improve correlations with 
human judgments. They can 
be used as concise parameterizations of PPDB, thereby vastly increasing its coverage. 
We also perform a qualitative analysis of the differences between additive and RNN composition.

\vspace{2pt}
\noindent\textbf{Introduce two new datasets}. The first contains PPDB phrase pairs and evaluates how well models can measure the quality of short paraphrases. 
The second is a new annotation of the bigram similarity task in \newcite{Mitchell:Lapata:2010} that makes it suitable for evaluating bigram paraphrases.

We release the new datasets, complete with annotation instructions and raw annotations, as well as our code and the trained models.\footnote{available on the authors' websites}

\section{Related Work}
There is a vast literature on representing words as vectors. The intuition of 
most methods to create these vectors (or embeddings) is that similar words have similar contexts~\cite{firth1957synopsis}. Earlier models made use of latent semantic analysis (LSA) \cite{deerwester1990indexing}. Recently, more sophisticated neural models, work originating with \cite{bengio2003neural}, have been gaining popularity \cite{mikolov2013efficient,pennington2014glove}. These embeddings are now being used in new ways as they are being tailored to specific downstream tasks \cite{bansal2014tailoring}.

\emph{Phrase} representations can be created from word vectors using compositional models. 
Simple but effective compositional models were 
studied by Mitchell and Lapata~\shortcite{mitchell2008vector,Mitchell:Lapata:2010} and \newcite{Blacoe2012}. They compared a variety of binary operations on word vectors and found that simple point-wise multiplication of 
explicit vector representations performed very well.  Other works like \newcite{zanzotto2010estimating} and \newcite{baroni2010nouns} also explored composition using models based on operations of vectors and matrices. 

More recent work has shown that the extremely efficient neural embeddings of \newcite{mikolov2013efficient} also do well on compositional tasks simply by adding the word vectors \cite{mikolov2013distributed}. \newcite{hashimoto2014} introduced an alternative word embedding and compositional model based on predicate-argument structures that does well on two simple composition tasks, including the one introduced by \newcite{Mitchell:Lapata:2010}.

An alternative approach to composition, used by \newcite{SocherEtAl2011:PoolRAE}, is to train a recursive neural network (RNN) whose structure is defined by a binarized parse tree. In particular, they trained their RNN as an unsupervised autoencoder. 
The RNN captures the latent structure of composition.
Recent work has shown that this model struggles 
in tasks involving compositionality~\cite{Blacoe2012,hashimoto2014}.\footnote{We also replicated this approach 
and found training to be time-consuming 
even using low-dimensional word vectors.}
However, we 
found success using RNNs in a \emph{supervised} setting, similar to \newcite{SocherKLMN14}, who used RNNs
to learn representations for image descriptions. The objective function we used in this work was motivated by their multimodal objective function for learning joint image-sentence representations.

Lastly, the PPDB has been used along with other resources to learn word embeddings for several tasks, including semantic similarity, language modeling, predicting human judgments, and classification~\cite{yu-14,faruqui-15}. Concurrently with our work, it has also been used to construct paraphrase models for short phrases \cite{TACL586}.

\section{New Paraphrase Datasets}
We created two novel datasets: (1) \annoppdb, a subset of phrase pairs from PPDB which 
are annotated according to how strongly they represent a paraphrase relationship, and (2) \mlpara, a re-annotation of the bigram similarity dataset from \newcite{Mitchell:Lapata:2010}, again annotated for strength of paraphrase relationship. 

\subsection{\annoppdb}
\label{sec:ppdb-task}

Our motivation for creating \annoppdb was to establish a way to 
evaluate compositional paraphrase models 
on \emph{short phrases}. 
Most existing paraphrase tasks focus on words, like SimLex-999~\cite{HillRK14}, or entire sentences, such as the Microsoft Research Paraphrase Corpus~\cite{dolan-04,quirk-04}. To our knowledge, there are no datasets that focus on the paraphrasability of short phrases. Thus, we created \annoppdb so that researchers can focus 
on local compositional phenomena and measure the performance of 
models directly\----avoiding the need to do so indirectly in a sentence-level task. Models that have strong performance on \annoppdb can be used to provide more accurate confidence scores for the paraphrases in the PPDB as well as reduce the need for large paraphrase tables altogether.

\annoppdb was created in a multi-step process 
(outlined below) involving various automatic filtering steps followed by crowdsourced human annotation. One of the aims for our dataset was to collect a variety of paraphrase types\----we wanted to include pairs that were non-trivial to recognize as well as those with a range of similarity and length. We focused on 
phrase pairs with limited lexical overlap to avoid including those with only trivial differences. 

We started with candidate phrases extracted from the first 10M pairs in the XXL version of the PPDB and then executed the following steps.\footnote{Note that the confidence scores for phrase pairs in PPDB are based on a weighted combination of features with weights determined heuristically. The confidence scores were used to place the phrase pairs into their respective sets (S, M, L, XL, XXL, etc.), where each larger set subsumes all smaller ones.}

\noindent\textbf{\bf Filter phrases for quality:}  Only those phrases whose tokens were in our vocabulary were retained.\footnote{Throughout, our vocabulary is defined as the most common 100K word types in English Wikipedia, following tokenization and lowercasing (see \secref{sec:word}).} Next, all duplicate paraphrase pairs were removed; in PPDB, these are distinct pairs that contain the same two phrases with the order swapped.

\noindent\textbf{\bf Filter by lexical overlap:}  Next, we calculated the {\it word overlap score} in each phrase pair and then retained only those pairs that had a score of less than $0.5$. By {\it word overlap score}, we mean the fraction of tokens in the smaller of the phrases 
with Levenshtein distance $\leq 1$ 
to a token in the larger of the phrases. This was done to exclude less 
interesting phrase pairs like $\langle${\it my dad had}, {\it my father had}$\rangle$ or $\langle${\it ballistic missiles}, {\it of ballistic missiles}$\rangle$ that only differ in a synonym or the addition of a single word.

\noindent\textbf{\bf Select range of paraphrasabilities:} To balance our dataset with both clear paraphrases and erroneous pairs in PPDB, we sampled 5,000 examples from ten chunks of the first 10M initial phrase pairs where a chunk is defined as 1M phrase pairs. 

\noindent\textbf{\bf Select range of phrase lengths:} We then 
selected 1,500 phrases from each 5000-example sample that encompassed a wide range of phrase lengths. To do this, we first binned the phrase pairs by their {\it effective size}. Let $n_1$ be the number of tokens of length greater than one character in the first phrase and $n_2$ the same for the second phrase. Then the {\it effective size} is defined as $\max(n_1,n_2)$. 
The bins contained pairs of {\it effective size}  of 3, 4, and 5 or more, and 500 pairs were selected from each bin. This gave us a total of 15,000 phrase pairs.

\noindent\textbf{\bf Prune to 3,000:} 3,000 phrase pairs were then selected randomly from the 15,000 remaining pairs to form an initial dataset, \annoppdbthreek. The phrases were selected so that every phrase in the dataset was unique. 

\noindent\textbf{\bf Annotate with Mechanical Turk:} The dataset was then rated on a scale from 1-5 using Amazon Mechanical Turk, where a score of 5 denoted phrases that are equivalent in a large number of contexts, 3 meant that the phrases had some overlap in meaning, and 1 indicated that the phrases were dissimilar or contradictory in some way (e.g., {\it can not adopt} and {\it is able to accept}).

We only permitted workers whose location was in the United States and who had done at least 1,000 HITS with a 99\% acceptance rate. Each example was labeled by 5 annotators and their scores were averaged to produce the final rating. Table~\ref{table:ppdbdata} shows some statistics of the data. Overall, the annotated data had a mean deviation (MD)\footnote{MD is similar to standard deviation, but uses absolute value instead of squared value and thus is both more intuitive and less sensitive to outliers.} of 0.80. Table~\ref{table:ppdbdata} shows that overall, workers found the phrases to be of high quality, as more than two-thirds of the pairs had an average score of at least 3. Also from the Table, we can see that workers had stronger agreement on very low and very high quality pairs and were less certain in the middle of the range.

\noindent\textbf{\bf Prune to 1,260:} To create our final dataset, \annoppdb, we selected 1,260 phrase pairs from the 3,000 annotations. We did this by first binning the phrases into 3 categories: those with scores in the interval $[1,2.5)$, those with scores in the interval $[2.5,3.5]$, and those with scores in the interval $(3.5,5]$. We took the 420 phrase pairs with the lowest MD in each bin, as these have the most agreement about their label, to form \annoppdb.

These 1,260 examples were then randomly split into a development set of 260 examples and a test set of 1,000 examples. The development set had an MD of 0.61 and the test set had an MD of 0.60, indicating the final dataset had pairs of higher agreement than the initial 3,000.

\begin{table}[t]
\centering
\small
\begin{tabular}{ | l || c |  c |} 
 \hline
\T Score Range & MD & \% of Data \\\hline
$[1,2)$ & 0.66 & 8.1\\
$[2,3)$ & 1.05 & 20.0\\
$[3,4)$ & 0.93 & 34.9\\
$[4,5]$ & 0.59 & 36.9\\
\hline
\end{tabular}
\caption{An analysis of \annoppdbthreek extracted from PPDB. The statistics shown are for the splits of the data according to the average score by workers. MD denotes mean deviation and \% of Data refers to the percentage of our dataset that fell into each range.\vspace{-0.4cm}}
\label{table:ppdbdata}
\end{table}

\subsection{\mlpara}
\label{sec:bigram-task}

Our second newly-annotated dataset, \mlpara, is based on the bigram similarity task originally introduced by \newcite{Mitchell:Lapata:2010}; we refer to the original annotations as the ML dataset. 

The ML dataset consists of human similarity ratings for three types of bigrams: adjective-noun (JN), noun-noun (NN), and verb-noun (VN). 
Through manual inspection, we found that the annotations were not consistent with the notion of similarity central to paraphrase tasks. For instance, {\it television set} and {\it television programme} were the highest rated phrases in the NN section (based on average annotator score). Similarly, one of the highest ranked JN pairs was {\it older man} and {\it elderly woman}. This indicates that the annotations reflect topical similarity in addition to capturing functional or definitional similarity.

Therefore, we had the data re-annotated by two authors of this paper who are native English speakers.\footnote{We tried using Mechanical Turk here, but due to such short phrases, with few having the paraphrase relationship, workers did not perform well on the task.}
The bigrams were labeled on a scale from 1-5 where 5 denotes phrases that are equivalent in a large number of contexts, 
3 indicates the phrases are roughly equivalent in a narrow set of contexts, and 
1 means the phrases are not at all equivalent in any context. Following annotation, we collapsed the rating scale by merging 4s and 5s together and 1s and 2s together. 

\begin{table}[h]
\centering
\scriptsize
\small
\begin{tabular}{ | l || c | c | c | c |} 
 \hline
\T Data & IA $\rho$ & IA $\kappa$ & ML comp. $\rho$ & ML Human $\rho$ \\\hline
JN & 0.87 & 0.79 & 0.56 & 0.52\\
NN & 0.64 & 0.58 & 0.38 & 0.49\\
VN & 0.73 & 0.73 & 0.55 & 0.55\\
\hline
\end{tabular}
\caption{
Inter-annotator agreement of \mlpara and comparison with ML dataset. Columns 2 and 3 show the inter-annotator agreement between the two annotators measured with Spearman $\rho$ and Cohen's $\kappa$.  Column 4 shows the $\rho$ between \mlpara and all of the ML dataset. The last column is the average human $\rho$ on the ML dataset. 
\vspace{-0.4cm}}
\label{table:bigramdata}
\end{table}

Statistics for the data are shown in Table~\ref{table:bigramdata}.
We show 
inter-annotator Spearman $\rho$ and Cohen's $\kappa$ in columns 2 and 3, indicating substantial agreement on the JN and VN portions but only moderate agreement on NN. 
In fact, when evaluating our NN annotations 
against those 
from the original ML data (column 4), we find $\rho$ to be 0.38, well below the average human correlation of 0.49 (final column)
reported by Mitchell and Lapata
and also 
surpassed by 
pointwise multiplication \cite{Mitchell:Lapata:2010}. 
This suggests that the original NN portion, more so than the others, favored a notion of similarity more related to association than paraphrase.

\section{Paraphrase Models}

We now present parametric paraphrase models and discuss training. Our goal is to
embed phrases into a low-dimensional space such that cosine similarity in the space corresponds to the strength of the paraphrase relationship between phrases.

We use 
a recursive neural network (RNN) 
similar to that used by 
\newcite{SocherKLMN14}. 
We first use a constituent parser to obtain a binarized parse of a phrase. For phrase $p$, we compute its vector $g(p)$ through recursive computation on the parse. That is, if phrase $p$ is the yield of a parent  node in a parse tree, and phrases $c_1$ and $c_2$ are the yields of its two child nodes, we define $g(p)$ recursively as follows:
\begin{equation}
g(p) = f(W[g(c_1);g(c_2)] + b)\nonumber
\end{equation}
\noindent where $f$ is an element-wise activation function ($\tanh$), $[g(c_1);g(c_2)] \in \mathbb{R}^{2n}$ is the concatenation of the child vectors, $W \in \mathbb{R}^{n \times 2n}$ is the composition matrix, $b \in \mathbb{R}^{n}$ is the offset, and $n$ is the dimensionality of the word embeddings. 
If node $p$ has no children (i.e., it is a single token), we define $g(p)=W_w^{(p)}$, where $W_w$ is the word embedding matrix in which particular word vectors are indexed using superscripts. 
The trainable parameters of the model are $W$, $b$, and $W_w$.

\subsection{Objective Functions}

We now present objective functions for training on 
pairs extracted from PPDB. The training data consists of (possibly noisy) pairs taken directly from the original PPDB. In subsequent sections, we discuss how we extract training pairs for particular tasks.

We assume our training data consists of a set $X$ of phrase pairs $\langle x_1, x_2\rangle$, where $x_1$ and $x_2$ are assumed to be paraphrases. To learn the model parameters ($W,b,W_w$), we minimize our objective function over the data using AdaGrad~\cite{Duchi} with mini-batches. The objective function follows:
\begin{multline} \label{eq:phrase}
\underset{W,b,W_w}{\text{min}} \frac{1}{|X|}\Bigg(\sum_{\langle x_1,x_2\rangle \in X} \\
\max(0,\delta - g(x_1)\cdot g(x_2) + g(x_1) \cdot g(t_1)) \\
+ \max(0,\delta - g(x_1)\cdot g(x_2) + g(x_2) \cdot g(t_2))\bigg) \\
+ \lambda_W (\norm{W}^2 + \norm{b}^2) + \lambda_{W_w}\norm{W_{w_{\mathit{initial}}} - W_w}^2
\end{multline} 
\noindent where $\lambda_{W}$ and $\lambda_{W_w}$ are regularization parameters, $W_{w_{\mathit{initial}}}$ is the initial word embedding matrix, $\delta$ is the \textbf{margin} (set to $1$ in all of our experiments), and $t_1$ and $t_2$ are carefully-selected \textbf{negative examples} taken from a mini-batch 
during optimization. 

The intuition for this objective is that we want the two phrases to be more similar to each other ($g(x_1)\cdot g(x_2)$) than either is to their respective negative examples $t_1$ and $t_2$, by a margin of at least $\delta$. 

\paragraph{Selecting Negative Examples}
\label{sec:sample}
To select 
$t_1$ and $t_2$ in Eq.~\ref{eq:phrase}, we simply chose the most similar phrase 
in the mini-batch (other than those in the given phrase pair). E.g., for choosing $t_1$ for a given $\langle x_1, x_2\rangle$:
\begin{equation}
t_1 = \argmax_{t : \langle t, \cdot\rangle \in X_b \setminus \{\langle x_1, x_2\rangle\}} g(x_1)\cdot g(t) \nonumber
\end{equation}
\noindent where $X_b\subseteq X$ is the current mini-batch. 
That is, 
we want to choose a negative example $t_i$ that is similar to $x_i$ according to the current model parameters. 
The downside of this approach is that we may occasionally choose a phrase $t_i$ that is actually a true paraphrase of $x_i$. 
We also tried a strategy in which we selected the least similar phrase that would trigger an update (i.e., $g(t_i)\cdot g(x_i) > g(x_1) \cdot g(x_2) - \delta$), but we found the simpler strategy above to work better and used it for all experiments reported below.

\paragraph{Discussion} 

The objective in Eq.~\ref{eq:phrase} is similar to one used by \newcite{SocherKLMN14}, but with several differences.
Their objective compared text and projected images. They also did not update the underlying word embeddings; we do so here, and in a way such that they are penalized from deviating from their initialization. Also for a given $\langle x_1, x_2\rangle$, they do not select a single $t_1$ and $t_2$ as we do, but use the entire training set, which can be very expensive with a large training dataset. 

We also experimented with a simpler objective that sought to directly minimize the squared L2-norm between $g(x_1)$ and $g(x_2)$ in each pair, along with the same regularization terms as in Eq.~\ref{eq:phrase}. One problem with this objective function is that the global minimum is $0$ and is achieved simply by driving the parameters to $0$. We obtained much better results using the objective in Eq.~\ref{eq:phrase}.

\paragraph{Training Word Paraphrase Models}
To train just word vectors on word paraphrase pairs (again from PPDB), we used the same objective function as above, but simply dropped the composition terms. This gave us an objective that bears some similarity to the \skipgram objective with negative sampling in \texttt{word2vec}~\cite{mikolov2013efficient}. Both seek to maximize the dot products of certain word pairs while minimizing the dot products of others. 
This objective function is:
\begin{multline} \label{eq:word}
\underset{W_w}{\text{min}} \frac{1}{|X|}\Bigg(\sum_{\langle x_1,x_2\rangle\in X} \text{max}(0,\delta - W_w^{(x_1)}\cdot W_w^{(x_2)} \\
+ W_w^{(x_1)} \cdot W_w^{(t_1)}) + \text{max}(0,\delta - W_w^{(x_1)}\cdot W_w^{(x_2)} + \\
W_w^{(x_2)} \cdot W_w^{(t_2)})\bigg)  + \lambda_{W_w}\norm{W_{w_{\mathit{initial}}} - W_w}^2
\end{multline}
\noindent It is like Eq.~\ref{eq:phrase} except with word vectors replacing the RNN composition function and with the regularization terms on the $W$ and $b$ removed. 

We further found we could improve 
this model by incorporating 
constraints. From our training pairs, for a given word $w$, we assembled all other words that were paired with it 
in PPDB and all of their lemmas. 
These were then used as constraints during the pairing process: a word $t$ could only be paired with $w$ if it was not in its list of assembled words. 

\section{Experiments -- Word Paraphrasing} 
\label{sec:word}

We first present experiments on learning lexical paraphrasability. We train on word pairs from PPDB and evaluate on the SimLex-999 dataset~\cite{HillRK14}, achieving the best results reported to date. 

\subsection{Training Procedure}
To learn word vectors that reflect paraphrasability, we optimized Eq.~\ref{eq:word}. 
There are many tunable hyperparameters with this objective, so to make training tractable we fixed the initial learning rates for the word embeddings to 0.5 and the margin $\delta$ to 1. Then we did a coarse grid search over a parameter space for $\lambda_{W_w}$ and the mini-batch size.
We considered $\lambda_{W_w}$ values in $\{10^{-2}, 10^{-3}, ..., 10^{-7}, 0\}$ and mini-batch sizes in \{100, 250, 500, 1000\}. We trained for 20 epochs for each set of hyperparameters using AdaGrad~\cite{Duchi}. 

For all experiments, we initialized our word vectors with \skipgram vectors trained using \texttt{word2vec}~\cite{mikolov2013efficient}. The vectors were trained on English Wikipedia (tokenized and lowercased, yielding 1.8B tokens).\footnote{We used the December 2, 2013 snapshot.
}  
We used a window size of 5  
and a minimum count cut-off of 60, 
producing vectors for approximately 270K word types. We retained vectors for only the 100K most frequent words, averaging the rest to obtain a single vector for unknown words. We will refer to this set of the 100K most frequent words as our \textbf{vocabulary}. 

\subsection{Extracting Training Data}

For training, we extracted word pairs from the lexical XL section of PPDB. We used the XL data for all experiments, including those for phrases. 
We used XL instead of XXL because XL has better quality overall while still being large enough so 
that we could be selective in choosing training pairs. 
There are a total of 548,085 pairs. We removed 174,766 that either contained numerical digits or words not in our 
vocabulary. 
We then removed 260,425 redundant pairs, 
leaving us with a final training set of 112,894 word pairs. 

\subsection{Tuning and Evaluation}
\label{sec:wordtune}

Hyperparameters were tuned using the wordsim-353 (\wsall) dataset~\cite{finkelstein2001placing}, specifically its 
similarity (\wssim) and relatedness (\wsrel) partitions~\cite{agirre2009study}. 
In particular, we tuned to maximize $2\times$\wssim correlation minus the \wsrel correlation. The idea 
was to reward vectors with high similarity 
and relatively low relatedness, in order to target the paraphrase relationship.

After tuning, we evaluated the best hyperparameters on the SimLex-999 (\simlex) dataset~\cite{HillRK14}.
We chose \simlex as our primary test set as it most closely evaluates the paraphrase relationship. 
Even though \wssim 
is a close approximation to this 
relationship, it 
does not include pairs that are merely associated and assigned low scores, 
which \simlex does (see discussion 
in Hill et al., 2014b\nocite{HillRK14}).

Note that for all experiments we used cosine similarity as our similarity metric and evaluated the statistical significance of dependent correlations using the one-tailed method of \cite{steiger1980tests}.

\subsection{Results}

\begin{table}[t]
\centering
\small
\begin{tabular} {| l |c|| c |} 
\hline
\T Model & $n$ & \simlex $\rho$ \\\hline
\skipgram & 25 & 0.21$\phantom{^{\ast}}$\\
\skipgram & 1000 & 0.38$\phantom{^{\ast}}$\\
\hline
\paragram$_{\mathrm{WS}}$ & 25 & 0.56$^{\ast}$\\
$\:\:\:$+ constraints & 25 &  \bf{0.58}$^{\ast}$\\
\hline
\newcite{HillRK14} & 200 &  0.446 \\
\newcite{hill-14} & - & 0.52$\phantom{^{\ast}}$ \\
inter-annotator agreement & N/A & 0.67$\phantom{^{\ast}}$ \\
\hline
\end{tabular}
\caption{
Results on the SimLex-999 (\simlex) word similarity task obtained by performing hyperparameter tuning based on $2\times$\wssim$ - $\wsrel and treating \simlex as a held-out test set. $n$ is word vector dimensionality. 
A $\ast$ indicates statistical significance ($p < 0.05$) over the 1000-dimensional skip-gram vectors. 
\vspace{-0.3cm}}
\label{fig:wordsim}
\end{table}

Table~\ref{fig:wordsim} shows results on \simlex when improving the initial word vectors by training on word pairs from PPDB, both with and without constraints. 
The ``\paragram$_{\mathrm{WS}}$'' rows show results when tuning to maximize $2\times$\wssim$ - $ \wsrel. 
We also show results for strong \skipgram baselines and the best results from the literature, including the state-of-the-art results from \newcite{hill-14} as well as the inter-annotator agreement 
from \newcite{HillRK14}.\footnote{\newcite{hill-14} did not report the dimensionality of the vectors that led to their state-of-the-art results.} 

The table illustrates that, by training on PPDB, we can surpass the previous best correlations on \simlex by 4-6\% absolute, 
achieving the best results reported to date.
We also find that 
we can train low-dimensional word vectors that exceed the performance of much larger vectors. 
This is very useful as using large vectors can 
increase both time and memory consumption 
in NLP applications.

To generate word vectors to use for downstream applications, we chose hyperparameters so as to maximize performance on \simlex.\footnote{We did not use constraints during training.} 
These word vectors, which we refer to as {\bf \paragram} vectors, had a $\rho$ of 0.57 on \simlex. We use them as initial word vectors for the remainder of the paper.

\subsection{Sentiment Analysis}
As an extrinsic evaluation of our \paragram word vectors, we used them in a convolutional neural network (CNN) for sentiment analysis. We used the simple CNN from \newcite{kim-14} and the binary sentence-level sentiment analysis task from \newcite{socher-13}. We used the standard data splits, removing examples with a neutral rating. 
We trained on all constituents in the training set while only using full sentences from development and test, giving us train/development/test sizes of 67,349/872/1,821. 

The CNN uses 
$m$-gram filters, each of which is an $m\times n$ vector. 
The CNN computes the inner product between an $m$-gram filter and each $m$-gram in an example, retaining the 
maximum match (so-called ``max-pooling''). The score of the match is a single dimension in a feature vector for the example, which is then associated with a weight 
in a linear classifier used to predict positive or negative sentiment. 

While \newcite{kim-14} used $m$-gram filters of several lengths, we only used unigram filters. We also fixed the word vectors during learning (called ``static'' by Kim). After learning, the unigram filters correspond to locations in the fixed word vector space. 
The learned classifier weights 
represent how strongly each location corresponds to positive or negative sentiment. We expect this static CNN to be more effective if the word vector space separates positive and negative sentiment. 

In our experiments, we compared baseline \skipgram embeddings to our \paragram vectors. 
We used AdaGrad learning rate of 0.1, mini-batches of size 10, 
and a dropout rate of 0.5. We used 200 unigram filters and rectified linear units as the activation (applied to the filter output $+$ filter bias). We trained for 30 epochs, 
predicting labels on the development set after each set of 3,000 examples. 
We recorded the highest development accuracy and used those parameters to predict labels on the test set. 

\begin{table}[t]
\centering
\small
\begin{tabular} {|l|c|c|}
\hline
word vectors & $n$ & accuracy (\%) \\ 
\hline
\skipgram & 25 & 77.0 \\
\skipgram & 50 & 79.6 \\\hline
\paragram & 25 & \textbf{80.9} \\
\hline
\end{tabular}
\caption{\label{tab:sentiment}Test set accuracies when comparing embeddings in a static CNN on the binary sentiment analysis task from \protect\newcite{socher-13}. 
\vspace{-0.4cm}}
\end{table}

Results are shown in Table~\ref{tab:sentiment}. 
We see 
improvements over the baselines when using \paragram vectors, even exceeding the performance of higher-dimensional \skipgram vectors.

\section{Experiments -- Compositional Paraphrasing}
In this section, we describe experiments on a variety of compositional phrase-based paraphrasing tasks. We start with the simplest case of bigrams, and then proceed to short phrases.
For all tasks, we again train on appropriate data from PPDB and test on various evaluation datasets, including our two novel datasets (\annoppdb and \mlpara).

\subsection{Training Procedure}
We trained our models by optimizing Eq.~\ref{eq:phrase} using AdaGrad~\cite{Duchi}. We fixed the initial learning rates to 0.5 for the word embeddings and 0.05 for the composition parameters, and the margin to 1. Then we did a coarse grid search over a parameter space for $\lambda_{W_w}$, $\lambda_{W}$, and mini-batch size.

For $\lambda_{W_w}$, our search space again consisted of $\{10^{-2}, 10^{-3}, ..., 10^{-7}, 0\}$, for $\lambda_W$ it was $\{10^{-1}, 10^{-2}, 10^{-3}, 0\}$, and we explored batch sizes of \{100, 250, 500, 1000, 2000\}. When initializing with \paragram vectors, the search space for $\lambda_{W_w}$ was shifted upwards to be $\{10, 1, 10^{-1}, 10^{-3}, ..., 10^{-6}\}$ to reflect our increased confidence in the initial vectors. We trained only for 5 epochs for each set of parameters.
For baselines, we used the same initial \skipgram vectors as in Section~\ref{sec:word}. 

\subsection{Evaluation and Baselines}
For all experiments, we again used cosine similarity as our similarity metric and evaluated the statistical significance using the method of \cite{steiger1980tests}. 

A baseline used in all compositional experiments is vector addition of \skipgram (or \paragram) 
word vectors. Unlike explicit word vectors, where point-wise multiplication acts as a conjunction of features and performs well on composition tasks \cite{mitchell2008vector}, using addition with \skipgram vectors \cite{mikolov2013distributed} gives better performance than multiplication.

\subsection{Bigram Paraphrasability}

\begin{table*}[th]
\setlength{\tabcolsep}{4pt} 
\small
\centering
\begin{tabular} { | lcc || >{\centering\arraybackslash}m{1cm} | >{\centering\arraybackslash}m{1cm} | >{\centering\arraybackslash}m{1cm} | >{\centering\arraybackslash}m{1cm} || c | c | c | c |} \hline
\multicolumn{3}{|c||}{Model} & 
      \multicolumn{4}{c||}{\newcite{Mitchell:Lapata:2010} Bigrams} &
      \multicolumn{4}{c|}{\mlpara} \\
word vectors & $n$ & comp. & JN & NN & VN & Avg & JN & NN & VN & Avg \\
\hline
\skipgram & 25 & + & 0.36\doubleast & 0.44\singleast  & 0.36\doubleast & 0.39 & 0.32\doubleast & 0.35\singleast  & 0.42\doubleast & 0.36 \\
\paragram & 25 & + &  0.44$^*$\singleast & 0.34\singleast  & 0.48$^*$\singleast  & 0.42 & 0.50$^*$\singleast  & 0.29\singleast & {\bf 0.58}$^{*\ddagger}$ & 0.46\\
\paragram & 25 & RNN  & {\bf 0.51}$^{*\dagger}$ & 0.40$^\dagger$ & \bf 0.50$^{*\ddagger}$ & \bf 0.47 & {\bf 0.57}$^{*\ddagger}$ & {\bf 0.44}$^\dagger$ & 0.55$^*$\singleast & \bf 0.52\\
\hline
\hline
\multicolumn{3}{|l||}{\newcite{hashimoto2014}} & 0.49\doubleast & 0.45\singleast  & 0.46\doubleast & {\bf 0.47} & 0.38\doubleast & 0.39\singleast  & 0.45\doubleast & 0.41 \\
\multicolumn{3}{|l||}{\newcite{Mitchell:Lapata:2010}}  & 0.46\doubleast & \bf 0.49\singleast & 0.38\doubleast & 0.44 & - & - & - & -\\
\hline
\multicolumn{3}{|l||}{Human} & - & - & -  & - & 0.87\doubleast & 0.64\singleast & 0.73\doubleast & 0.75 \\
\hline
\end{tabular}

\caption{
Results on the test section of the bigram similarity task of \protect\newcite{Mitchell:Lapata:2010} and our newly annotated version (\mlpara). 
 ($n$) shows the word vector dimensionality and (``comp.'') shows the composition function used: ``+'' is vector addition and ``RNN'' is the recursive neural network. The * indicates statistically significant ($p < 0.05$) over the \skipgram model, $\dagger$ statistically significant over the \{\paragram, +\} model, and $\ddagger$ statistically significant over \protect\newcite{hashimoto2014}. 
\vspace{-0.4cm}}
\label{fig:Bigram}
\end{table*}

To evaluate our ability to paraphrase bigrams, we consider the original bigram similarity task from \newcite{Mitchell:Lapata:2010} as well as our newly-annotated version of it: \mlpara. 

\paragraph{Extracting Training Data}

Training data for these tasks was extracted from the XL portion of PPDB. 
The bigram similarity task from \newcite{Mitchell:Lapata:2010} contains three types of bigrams: adjective-noun (JN), noun-noun (NN), and verb-noun (VN).  
We aimed to collect pairs from PPDB that mirrored these three types of bigrams. 

We found parsing to be unreliable on such short segments of text, so we used a POS tagger \cite{manning-EtAl:2014:P14-5} to tag the tokens in each phrase. We then used the word alignments in PPDB to extract bigrams for training. 
For JN and NN, we extracted pairs containing aligned, adjacent tokens in the two phrases with the appropriate part-of-speech tag. Thus we extracted pairs like $\langle${\it easy job}, {\it simple task}$\rangle$ for the JN section and $\langle${\it town meeting}, {\it town council}$\rangle$ for the NN section. We used a different strategy for extracting training data for the VN subset: we took aligned VN tokens and took the closest noun after the verb. This was done to approximate the direct object that would have been ideally extracted with a dependency parse. An example from this section is $\langle${\it achieve goal}, {\it achieve aim}$\rangle$. 

We removed phrase pairs that (1) contained words not in our vocabulary, (2) were redundant with others, (3) contained brackets, 
or (4) had Levenshtein distance $\leq 1$. 
The final criterion helps to ensure that we train on phrase pairs with non-trivial differences. 
The final training data consisted of 133,997 JN pairs, 62,640 VN pairs and 35,601 NN pairs. 

\paragraph{Baselines}
In addition to RNN models, we report baselines that use vector addition as the composition function, both with our  \skipgram embeddings and \paragram embeddings from Section~\ref{sec:word}.

We also compare to several results from prior work. When doing so, we took their \emph{best} correlations for each data subset. That is, the JN and NN results from \newcite{Mitchell:Lapata:2010} use their multiplicative model and the VN results use their dilation model. From \newcite{hashimoto2014} we used their PAS-CLBLM $\text{Add}_{\text{l}}$ and PAS-CLBLM $\text{Add}_{\text{nl}}$ models. We note that their vector dimensionalities 
are larger than ours, 
using $n=2000$ and $50$ 
respectively.

\paragraph{Results}

Results are shown in Table~\ref{fig:Bigram}. We report results on the test portion of the original \newcite{Mitchell:Lapata:2010} dataset (ML) as well as the entirety of our newly-annotated dataset (\mlpara). RNN results on ML were tuned on the respective development sections and RNN results on \mlpara were tuned on the entire ML dataset. 

Our RNN model outperforms results from the literature on most sections in both datasets and its average correlations are among the highest.\footnote{The results obtained here differ from those reported in \newcite{hashimoto2014} as we scored their vectors with a newer Python implementation of Spearman $\rho$ that handles ties (Hashimoto, P.C.).} 
The one subset of the data that posed difficulty was the NN section of the ML dataset. We suspect this is due to the reasons discussed in Section~\ref{sec:bigram-task}; for our \mlpara dataset, by contrast, we do see gains on the NN section. 

We also outperform the strong baseline of adding 1000-dimensional \skipgram embeddings, a model with 40 times the number of parameters, on our \mlpara dataset.
This baseline had correlations of 0.45, 0.43, and 0.47 
on the JN, NN, and VN partitions, with an average of 0.45{\----}below the average $\rho$ of the RNN (0.52) and even the \{\paragram, +\} model (0.46).

Interestingly, the type of vectors used to initialize the RNN has a significant effect on performance. If we initialize using the 25-dimensional \skipgram vectors, the average $\rho$ on \mlpara drops to 0.43, below even the \{\paragram, +\} model. 

\subsection{Phrase Paraphrasability}
\label{sec:phrase}

In this section we show that by training a model based on filtered phrase pairs in PPDB, we can actually distinguish between quality paraphrases and poor paraphrases in PPDB better than the original heuristic scoring scheme from \newcite{GanitkevitchDC13}. 

\paragraph{Extracting Training Data}

As before, training data was extracted from the XL section of PPDB. 
Similar to the procedure to create our \annoppdb dataset, phrases were filtered such that only those with a {\it word overlap score} of less than 0.5 were kept. We also removed redundant phrases and phrases that contained tokens not in our vocabulary. The phrases were then binned according to their {\it effective size} and 20,000 examples were selected from bins of {\it effective sizes} of 3, 4, and more than 5, creating a training set of 60,000 examples. Care was taken to ensure that none of our training pairs was also present in our development and test sets.

\paragraph{Baselines} 
We compare our models with strong lexical baselines. The first, {\it strict word overlap}, is the percentage of words in the smaller phrase that are also in the larger phrase. We also include a version where the words are lemmatized prior to the calculation.

We also train a support vector regression model (epsilon-SVR) \cite{chang2011libsvm} on the 33 features that are included for each phrase pair in PPDB. We scaled the features such that each lies in the interval $[-1,1]$ and tuned the parameters using 5-fold cross validation on our dev set.\footnote{We tuned both parameters over $\{2^{-10}, 2^{-9}, ..., 2^{10}\}$.} We then trained on the entire dev set after finding the best performing $C$ and $\epsilon$ combination and evaluated on the test set of \annoppdb.

\paragraph{Results}
\label{sec:phraseevaluation}

\begin{table}[h]
\setlength{\tabcolsep}{4pt} 
\small
\centering
\begin{tabular} { | lcc || c |} \hline
\multicolumn{3}{|c||}{Model} & \multicolumn{1}{c|}{}\\
word vectors & $n$ & comp. & \annoppdb \\\hline
\skipgram & 25 & + & 0.20\tripleast \\
\paragram & 25 & + & 0.32$^*$\doubleast \\
\paragram & 25 & RNN & \bf{0.40}$^{*\dagger\ddagger}$ \\
\hline
\hline
\multicolumn{3}{|l||}{\newcite{GanitkevitchDC13}} &  0.25\tripleast  \\
\multicolumn{3}{|l||}{word overlap (strict)} & 0.26\tripleast   \\
\multicolumn{3}{|l||}{word overlap (lemmatized)} & 0.20\tripleast  \\
\multicolumn{3}{|l||}{PPDB+SVR} & 0.33\tripleast \\
\hline
\end{tabular}
\caption{\label{fig:PPDB}
Spearman correlation on \annoppdb. The * indicates statistically significant ($p < 0.05$) over the \skipgram model, the $\dagger$ indicates statistically significant over the \{\paragram, +\} model, and the $\ddagger$ indicates statistically significant over PPDB+SVR.
\vspace{-0.3cm}}
\end{table}

We evaluated on our \annoppdb dataset described in \secref{sec:ppdb-task}. Table~\ref{fig:PPDB} shows the Spearman correlations on the 1000-example test set. RNN models were tuned on the development set of 260 examples. All other methods had no hyperparameters and therefore required no tuning. 

We note that the confidence estimates from \newcite{GanitkevitchDC13} reach a $\rho$ of 0.25 on the test set, similar to the results of strict overlap. While 25-dimensional \skipgram embeddings only reach 0.20, we can improve this to 0.32 by fine-tuning them using PPDB (thereby obtaining our \paragram vectors). 
By using the \paragram vectors to initialize the RNN, we reach a correlation of 0.40, which is better than the PPDB confidence estimates by 15\% absolute. 

We again consider addition of 1000-dimensional \skipgram embeddings as a baseline, and they continue to perform strongly ($\rho=0.37$). The RNN initialized with \paragram vectors does reach a higher $\rho$ (0.40), but the difference is not statistically significant ($p=0.16$). Thus we can achieve similarly-strong results with far fewer parameters.

This task also illustrates the importance of initializing our RNN model with appropriate word embeddings. An RNN initialized with \skipgram vectors has a modest $\rho$ of 0.22, well below the $\rho$ of the RNN initialized with \paragram vectors. Clearly, initialization is important when optimizing non-convex objectives like ours, but it is noteworthy that our best results came from first improving the word vectors and then learning the composition model, rather than jointly learning both from scratch.

\section{Qualitative Analysis}

\begin{table}[h]
\centering
\small
\begin{tabular}{ | c || c |  c |} 
 \hline
\T Score Range & + & RNN \\\hline
$[1,2)$ & 2.35  & 2.08 \\
$[2,3)$ &1.56 & 1.38\\
$[3,4)$ & 0.87 & 0.85\\
$[4,5]$ & 0.43 & 0.47\\
\hline
\end{tabular}
\caption{Average absolute error of addition and RNN models on different ranges of gold scores.}
\label{table:qualperf}
\end{table}

\begin{table*}[th!]
\setlength{\tabcolsep}{4pt} 
\small
\centering
\begin{tabular} { | c || c | c || c | c || c | c | c |} \hline
Index & Phrase 1 & Phrase 2 & Length Ratio & Overlap Ratio & Gold & RNN & + \\\hline
1 & scheduled to be held in & that will take place in & 1.0 & 0.4 & 4.6 & 2.9 & \bf 4.4\\
2 & according to the paper , & the newspaper reported that & 0.8 & 0.5 & 4.6 & 2.8 & \bf 4.1\\
\hline
\hline
3 & at no cost to & without charge to & 0.75 & 1.0 & 4.8 & 3.1 & \bf 4.6\\
4 &  's surname & family name of & 0.67 & 1.0 & 4.4 & 2.8 & \bf 4.1\\
\hline
\hline
5 & could have an impact on & may influence & 0.4 & 0.5 & 4.6 & \bf 4.2 & 3.2\\
6 & to participate actively & to play an active role & 0.6 & 0.67 & 5.0 & \bf 4.8 & 4.0\\
\hline
\hline
7 & earliest opportunity & early as possible & 0.67 & 0.0 & 4.4 & \bf 4.3 & 2.9\\
8 & does not exceed & is no more than & 0.75 & 0.0 & 5.0 & \bf 4.8 & 3.5\\

\hline
\end{tabular}
\caption{
Illustrative phrase pairs from \annoppdb with gold similarity $> 4$. 
The last three columns show the gold similarity score, the similarity score of the RNN model, and the similarity score of vector addition. We note that addition performs better when the pairs have high length ratio (rows 1--2) or overlap ratio (rows 3--4) while the RNN does better when those values are low (rows 5--6 and 7--8 respectively). Boldface indicates smaller error compared to gold scores. 
\vspace{-0.4cm}}
\label{table:examples}
\end{table*}

We performed a qualitative analysis to 
uncover sources of error and determine differences between adding \paragram vectors and using an RNN initialized with them. 
To do so, we took the output of both systems on \annoppdb and mapped their cosine similarities to the interval $[1,5]$. We then computed their absolute error as 
compared to the gold ratings. 

Table~\ref{table:qualperf} shows how the average of these absolute errors 
changes with the magnitude of the gold ratings. 
The RNN performs better (has lower average absolute error) for less similar pairs. 
Vector addition only does better 
on the most similar pairs. This is presumably because the most positive pairs have high word overlap and so can be represented effectively with a simpler model.

To further investigate the differences between these models, we 
removed those pairs with gold scores 
in $[2,4]$, in order to focus on pairs with extreme scores. 
We identified two factors that distinguished the performance between the two models: 
length ratio and the amount of lexical overlap. We did not find evidence that non-compositional phrases, such as idioms, were a source of error as these were not found in \mlpara and only appear rarely in \annoppdb. 

We define length ratio  
as simply the number of tokens in the smaller phrase divided by the number of tokens in the larger phrase. Overlap ratio is the number of {\it equivalent tokens} in the phrase pair divided by the number of tokens in the smaller of the two phrases. {\it Equivalent tokens} are defined as tokens that are either exact matches or are paired up in the lexical portion of PPDB used to train the \paragram vectors.

Table~\ref{table:lengthandoverlap} shows how the performance of the models changes under different values of length ratio and overlap ratio.\footnote{The bin delimiters were chosen to be uniform over the range of output values of the length ratio ([0.4,1] with one outlier data point removed) and overlap ratio ([0,1]). 
} The values in this table are the percentage changes in absolute error when using the RNN over the \paragram vector addition model. So negative values indicate superior performance by the RNN. 

\begin{table}[h]
\centering
\small
\begin{tabular}{ | c || c | c | c |} 
 \hline
\T \bf Length Ratio & $[0,0.6]$ & $(0.6, 0.8]$ & $(0.8,1]$ \\\hline
Positive Examples & -22.4 & 10.0 & 35.5 \\
Negative Examples & -9.9 &  -11.1 & -12.2 \\
Both & -13.0 & -6.4 & -2.0\\
\hline
\hline
\T \bf Overlap Ratio & [0,$\frac{1}{3}$] &  ($\frac{1}{3}$,$\frac{2}{3}$] & ($\frac{2}{3}$,1] \\\hline
Positive Examples & -4.5 & 7.0 & 19.4  \\
Negative Examples & -11.3 & -7.5 & -15.0\\
Both & -10.6 & -5.3 & 0.0 \\
\hline
\end{tabular}
\caption{Comparison of the addition and RNN model on phrase pairs of different overlap and length ratios. The values in the table are the percent change in absolute error from the addition model to the RNN model. Negative examples are defined as pairs from \annoppdb whose gold score is less than 2 and positive examples are those with scores greater than 4. ``Both'' refers to both negative and positive examples.
 }
\label{table:lengthandoverlap}
\end{table}

A few trends emerge from this table. One is that as the length ratio increases (i.e., the phrase pairs are closer in length), addition surpasses the RNN for positive examples. For negative examples, the trend is reversed. 
The same trend appears for overlap ratio. 
Examples from \annoppdb illustrating these trends on positive examples are shown in Table~\ref{table:examples}. 

When considering both positive and negative examples (``Both''), we see that the RNN excels 
on the most difficult examples (large differences in phrase length and less lexical overlap). 
For easier examples, the two fare similarly overall (-2.0 to 0.0\% change), but the RNN does much better on negative examples. 
This aligns with the intuition that addition should perform well when two paraphrastic phrases have high lexical overlap and similar length. But when they are not paraphrases, simple addition is misled and the RNN's learned composition function better captures the relationship. This may suggest new architectures for modeling compositionality differently depending on differences in length and amount of overlap.

\section{Conclusion}
\begin{table*}
\centering
\small
\begin{tabular} {| l |c|| c | c | c | c |} 
\hline
\T Model & $n$ & \simlex & \wsall & \wssim & \wsrel \\\hline
GloVe & 300 & 0.376 & 0.579 & 0.630 & 0.571\\
\hline
\paragramsubs{\mathrm{300,\wsall}} & 300 & 0.667 & \bf 0.769 & \bf 0.814 & \bf 0.730 \\
\paragramsubs{\mathrm{300,\simlex}} & 300 & \bf 0.685 & 0.720 & 0.779 & 0.652  \\
\hline
inter-annotator agreement$^*$& N/A & 0.67 & 0.756 & N/A & N/A \\
\hline
\end{tabular}
\caption{Evaluation of 300 dimensional \paragram vectors on \simlex and \wsall. Note that the inter-annotator agreement $\rho$ was calculated differently for \wsall and \simlex. For \simlex,
the agreement was computed as the average pairwise correlation between pairs of annotators, while for \wsall, agreement was computed as the average correlation between a single annotator with
the average over all other annotators. If one uses the alternative measure of agreement for \wsall, the agreement is 0.611, which is easily beaten by automatic methods \protect\cite{HillRK14}.
\vspace{-0.3cm}}
\label{fig:wordsim2}
\end{table*}

\begin{table*}
\setlength{\tabcolsep}{4pt} 
\small
\centering
\begin{tabular} { | lcc || >{\centering\arraybackslash}m{1cm} | >{\centering\arraybackslash}m{1cm} | >{\centering\arraybackslash}m{1cm} | >{\centering\arraybackslash}m{1cm} || c | c | c | c |} \hline
\multicolumn{3}{|c||}{Model} & \multicolumn{4}{c||}{\newcite{Mitchell:Lapata:2010} Bigrams} & \multicolumn{4}{c|}{\mlpara}\\
word vectors & $n$ & comp. & JN & NN & VN & Avg & JN & NN & VN & Avg  \\
\hline
GloVe & 300 & + & 0.40 & 0.46 & 0.37 & 0.41 & 0.39 & 0.36 & 0.45 & 0.40 \\
\hline
\paragramsubs{\mathrm{300,\wsall}} & 300 & + & 0.52 & 0.41 & 0.49 & \bf 0.48 & 0.55 & 0.42 & 0.55 & 0.51 \\
\paragramsubs{\mathrm{300,\simlex}} & 300 & + & 0.51 & 0.36 & 0.51 & 0.46 & 0.57 & 0.39 & 0.59 & \bf 0.52 \\
\hline
\end{tabular}
\caption{
Evaluation of 300 dimensional \paragram vectors on the bigram tasks.
\vspace{-0.4cm}}
\label{fig:Bigram2}
\end{table*}

We have shown how to leverage PPDB to learn state-of-the-art word embeddings and compositional models for paraphrase tasks. 
Since PPDB was 
created automatically from parallel corpora, 
our models are also built automatically. 
Only small amounts of annotated data are used to tune hyperparameters.

We also introduced two new datasets to evaluate compositional models of short paraphrases, filling a gap in the NLP community, as currently there are no datasets created for this purpose. Successful models on these datasets can then be used to extend the coverage of, or provide an alternative to, PPDB.

There remains 
a great deal of work to be done in 
developing new composition models, 
whether with new network architectures or distance functions. 
In this work, we based our composition function on constituent parse trees, but this may not be the best approach\----especially for short phrases. Dependency syntax may be a better alternative~\cite{SocherKLMN14}.  
Besides improving composition, 
another direction 
to explore is how to 
use models for short phrases in sentence-level paraphrase recognition and other downstream tasks.

\begin{appendices}
\section{}
Increasing the dimension of word embeddings or training them on more data can have a significant positive impact on many tasks\----both at the word level and on downstream tasks.  We scaled up our original 25-dimensional \paragram embeddings and modified our training procedure slightly in order to produce two sets of 300-dimensional \paragram vectors.\footnote{Both \paragramsubs{\mathrm{300,\wsall}} and \paragramsubs{\mathrm{300,\simlex}} vectors can be found on the authors' websites.} The vectors outperform our original 25-dimensional \paragram vectors on all tasks and achieve human-level performance on \simlex and \wsall. Moreover, when simply using vector addition as a compositional model, they are both on par with the RNN models we trained specifically for each task. These results can be seen in Tables~\ref{fig:wordsim2},~\ref{fig:Bigram2}, and~\ref{fig:PPDB2}. 

\begin{table}
\setlength{\tabcolsep}{4pt} 
\footnotesize
\centering
\begin{tabular} { | lcc || c |} \hline
\multicolumn{3}{|c||}{Model} & \multicolumn{1}{c|}{}\\
word vectors & $n$ & comp. & \annoppdb \\
\hline
GloVe & 300 & + & 0.27 \\
\hline
\paragramsubs{\mathrm{300,\wsall}} & 300 & + & \bf 0.43 \\
\paragramsubs{\mathrm{300,\simlex}} & 300 & + & 0.41 \\
\hline
\end{tabular}
\caption{\label{fig:PPDB2}
Evaluation of 300 dimensional \paragram vectors on \annoppdb.
\vspace{-0.3cm}}
\end{table}

The main modification was to use higher-dimensional initial embeddings, in our case the pretrained 300-dimensional GloVe embeddings.\footnote{We used the GloVe vectors trained on 840 billion tokens of Common Crawl data, available at \url{http://nlp.stanford.edu/projects/glove/}} Since PPDB only contains lowercased words, we extracted only one GloVe vector per word type (regardless of case) by taking the first occurrence of each word in the vocabulary. This is the vector for the most common casing of the word, and was used as the word's single initial vector in our experiments. This reduced the vocabulary from the original 2.2 million types 
to 1.7 million.

Smaller changes included replacing dot product with cosine similarity in Equation~\ref{eq:word} and a change to the negative sampling procedure. We experimented with three approaches: {\it MAX sampling} discussed in Section~\ref{sec:sample}, {\it RAND sampling} which is random sampling from the batch, and 
a 50/50 mixture of {\it MAX sampling} and {\it RAND sampling}. 

For training data, we selected all word pairs in the lexical portion of PPDB XL that were in our vocabulary, removing redundancies. This resulted in 169,591 pairs for training. 
We trained our models for 10 epochs and tuned hyperparameters (batch size, $\lambda_{W_w}$, $\delta$, and sampling method) 
in two 
ways: maximum correlation on \wsall  (\paragramsubs{\mathrm{300,\wsall}}) and maximum correlation on \simlex (\paragramsubs{\mathrm{300,\simlex}}).\footnote{Note that if we use the approach in Section~\ref{sec:wordtune} in which we tune to maximize $2\times$\wssim correlation minus the \wsrel correlation, the \simlex $\rho$ is 0.640, still higher than any other reported result to the best of our knowledge.} We report results for both sets of embeddings in Tables~\ref{fig:wordsim2},~\ref{fig:Bigram2}, and~\ref{fig:PPDB2}, and make both available to the community in the hope that they may be useful for other downstream tasks.

\end{appendices}

\section*{Acknowledgements}
We thank the editor and the anonymous reviewers as well as Juri Ganitkevitch, Weiran Wang, and Kazuma Hashimoto for their valuable comments and technical assistance. 
We also thank 
Chris Callison-Burch, 
Dipanjan Das, 
Kuzman Ganchev, 
Ellie Pavlick, 
Slav Petrov, 
Owen Rambow, 
David Sontag, 
Oscar T\"{a}ckstr\"{o}m, 
Kapil Thadani, 
Lyle Ungar, 
Benjamin Van Durme, 
and Mo Yu 
for helpful conversations. 
This research was supported by a Google Faculty Research Award to Mohit Bansal, Karen Livescu, and Kevin Gimpel, the Multimodal Information Access \& Synthesis Center at UIUC, part of CCICADA, a DHS Science and Technology Center of Excellence, and by DARPA under agreement number FA8750-13-2-0008. The views and conclusions contained herein are those of the authors and should not be interpreted as necessarily representing the official policies or endorsements, either expressed or implied, of DARPA or the U.S. Government. 
\bibliographystyle{acl2012}
\bibliography{paraphrase}

\end{document}